\documentclass[final,5p,times,twocolumn]{elsarticle}

\usepackage{amssymb}
\usepackage{amsmath}

\usepackage{hyperref}
\usepackage[capitalise]{cleveref}
\crefname{appendix}{}{} 
\usepackage{booktabs}
\usepackage{siunitx}
\usepackage{multirow}
\usepackage{bm} 
\usepackage{dsfont} 

\usepackage{tikz}
\usetikzlibrary{colorbrewer}
\usetikzlibrary{calc}
\usetikzlibrary{patterns}
\usetikzlibrary{decorations}
\usetikzlibrary{external}
\usetikzlibrary{positioning}
\usepackage{pgfplots}
\input{tikz/turbo.colormap}  
\input{tikz/rainbow.colormap}  
\input{tikz/rainbow_uniform.colormap}  
\pgfplotsset{compat=1.16}
\usepgfplotslibrary{colorbrewer}
\usepgfplotslibrary{fillbetween}
\usepgfplotslibrary{groupplots}

\newcommand{\ppvec}[1]{\bm{#1}}
\newcommand{\ppmat}[1]{\underline{\bm{#1}}}

\journal{}

\begin{document}

\begin{frontmatter}



\title{Invariant Control Strategies for Active Flow Control using Graph Neural Networks}

\author[label1]{Marius Kurz\fnref{fn1}}
\author[label2]{Rohan Kaushik\fnref{fn1}\corref{cor1}}
\author[label2]{Marcel Blind}
\author[label2]{Patrick Kopper}
\author[label2]{Anna Schwarz}
\author[label2]{Felix Rodach}
\author[label2]{Andrea Beck}

\affiliation[label1]{organization={Centrum Wiskunde \& Informatica (CWI)},
            addressline={Science Park 123},
            city={Amsterdam},
            postcode={1098 XG},
            country={The Netherlands}}

\affiliation[label2]{organization={Institute of Aerodynamics and Gas Dynamics, University of Stuttgart},
            addressline={Wankelstraße 3},
            postcode={70563},
            city={Stuttgart},
            country={Germany}}

\fntext[fn1]{M. Kurz and R. Kaushik contributed equally and cordially agree to share first authorship.}
\cortext[cor1]{Corresponding author}


\begin{abstract}
Reinforcement learning (RL) has recently gained traction for active flow control tasks, with initial applications exploring drag mitigation via flow field augmentation around a two-dimensional cylinder.
RL has since been extended to more complex turbulent flows and has shown significant potential in learning complex control strategies. However, such applications remain computationally challenging owing to its sample inefficiency and associated simulation costs.
This fact is worsened by the lack of generalization capabilities of these trained policy networks, often being implicitly tied to the input configurations of their training conditions.
In this work, we propose the use of graph neural networks (GNNs) to address this particular limitation, effectively increasing the range of applicability and getting more \textit{value} out of the upfront RL training cost.
GNNs can naturally process unstructured, three-dimensional flow data, preserving spatial relationships without the constraints of a Cartesian grid.
Additionally, they incorporate rotational, reflectional, and permutation invariance into the learned control policies, thus improving generalization and thereby removing the shortcomings of commonly used convolutional neural networks (CNNs) or multilayer perceptron (MLP) architectures.
To demonstrate the effectiveness of this approach, we revisit the well-established two-dimensional cylinder benchmark problem for active flow control.
The RL training is implemented using Relexi, a high-performance RL framework, with flow simulations conducted in parallel using the high-order discontinuous Galerkin framework FLEXI.
Our results show that GNN-based control policies achieve comparable performance to existing methods while benefiting from improved generalization properties.
This work establishes GNNs as a promising architecture for RL-based flow control and highlights the capabilities of Relexi and FLEXI for large-scale RL applications in fluid dynamics.
\end{abstract}

%


\end{frontmatter}



\section{Introduction}

Reinforcement learning (RL) has gained increasing attention for the task of active flow control (AFC)~\cite{vignon2023recent,vinuesa2024perspectives}.
One of the first applications of RL for AFC was reported by \citet{rabault2019artificial}, who employed RL to control the flow around a two-dimensional cylinder using a pair of inflow and suction jets at the cylinder's poles.
Since then, this case has established itself as a well-known benchmark problem in the field, having been extensively studied in the literature \cite{rabault2019artificial,rabault2019accelerating,ren2021applying,varela2022deep,weiner2024modelbased} and serving as inspiration for a plethora of related problems (e.g.~\cite{jiang2023reinforcement,han2022deep}).
More recently, RL has been applied to the more complex task of controlling fully three-dimensional turbulent flow, for instance turbulent channel flow, to either reduce drag through suction and blowing at one of the walls~\cite{guastoni2023deep,sonoda2023reinforcement} or control the wall cycle using streamwise travelling waves~\cite{cavallazzi2024deep}.
Other applications entail the control of turbulent flow around a three-dimensional cylinder~\cite{suarez2025flow} or the control of a separation bubble in separated flow~\cite{font2025deep}.

One limiting factor for such applications is the computationally intensive nature of turbulent flow simulations, which poses significant challenges for the application of RL to these cases.
This has been somewhat addressed by frameworks leveraging the efficiency of established flow solvers and distributed computing resources to accelerate training~\cite{wang2022drlinfluids,shams2023gymprecice,kurz2022relexi}.
Unfortunately, RL is oftentimes found to be sample-inefficient in practice~\cite{jin2018qlearning}, thereby requiring a considerable amount of simulations to find a successful control policy.

It has been increasingly acknowledged that incorporating invariance and equivariance properties into the employed neural network architectures can be crucial to improving generalizability and training efficiency~ \cite{vignon2023recent}.
Despite the great potential of these approaches, only few studies have investigated their use in the context of flow control. 
One example was reported by \citet{vignon2023effective}, who demonstrated that a multi-agent reinforcement learning (MARL) approach with a translationally invariant control law improves the performance of RL-enabled control significantly for the three-dimensional Rayleigh-B{\'e}rnard problem.
Similarly, \citet{peitz2024distributed} showed that a distributed, translationally invariant control law can reduce the complexity of the control problem and improve the transferability across different problem sizes for a range of dynamical systems.
While they used a standard convolutional neural network (CNN) architecture, \citet{jeon2024advanced} employed a group-invariant CNN architecture for the Rayleigh-B{\'e}rnard problem, demonstrating improved performance and faster convergence of the RL agent.

Thus, in light of the highlighted compute cost and the limited generalization of traditional architectures, in this work we propose the use of graph neural networks (GNNs) for AFC to address these issues.
The improved generalizability of GNNs at comparable compute overhead effectively allows one to extract more value out of the upfront RL training cost.
GNNs can handle pointwise, unstructured data straight-forwardly.
This allows retaining the spatial structure of the flow field, while not restricting the probes to be distributed in a Cartesian grid - as would be required for the commonly used CNN architectures.
GNNs also allow embedding rotational and reflectional equivariance into the network architecture, which enables learning control laws that are invariant under these transformations.
Moreover, GNNs by design enforce permutation invariance into the control law, such that the control law becomes independent of the specific ordering of the input features.
Thus, we revisit the two-dimensional cylinder flow benchmark problem~\cite{rabault2019artificial}.
The RL training is implemented using Relexi~\cite{kurz2022relexi}, which is an efficient RL framework for high-performance computing~(HPC) that has already been applied to several applications in turbulence modeling~\cite{kurz2022deep,kurz2023deep,beck2023discretization}.
Here, the training environments are provided by running multiple flow simulations with the high-order accurate FLEXI solver~\cite{krais2021flexi} in parallel.
Within this framework, we apply GNNs to the task of learning invariant control laws for AFC.
To the best of our knowledge, this is the first work employing GNNs for an application in AFC.

This paper is organized as follows.
In \cref{sec:methodology}, we introduce the methodology, outlining the simulation setup, the network architecture, and the training procedure using RL.
In this work, we will learn control strategies for the described cylinder flow with both an MLP and a GNN as an agent. The MLP serves to validate our approach against existing literature, accounting for RL’s stochasticity in AFC, and provides a baseline for comparison with the GNN. Notably, MLPs lack the invariances and equivariances expected in GNNs.
The results of the training are reported in \cref{sec:results}, where the MLP performance is first validated against results from literature, followed by a performance comparison with the GNNs. Further demonstrated are the proposed invariance properties of the GNNs.
Finally, \cref{sec:conclusion} concludes the paper with a summary and discussion of its key findings.


\section{Methodology}
\label{sec:methodology}

\subsection{Simulation Environment}
\label{subsec:simulation}

\begin{figure*}[t]
    \centering
    \tikzsetnextfilename{fig_simulation_setup}
    \input{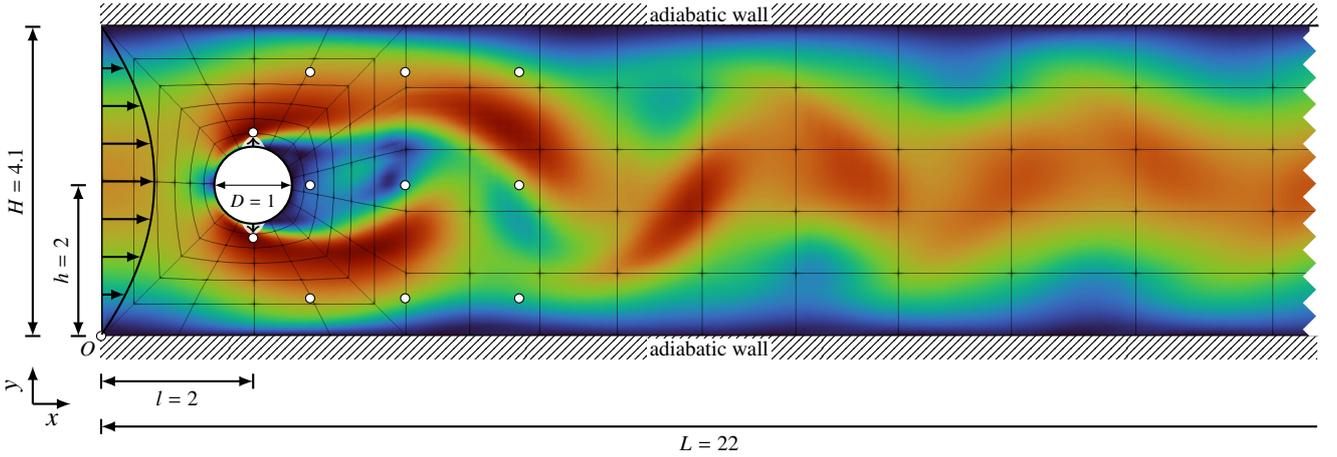}
    \caption{Simulation setup for the flow around a two-dimensional cylinder in a channel. The positions of the 11 pressure probes are highlighted by white circles and the jets by white areas at the top and bottom of the cylinder. The field solution shows the velocity magnitude. The domain is clipped here to a length of $y=16$ for better visual representation.}
    \label{fig:simulation:setup}
\end{figure*}

The simulation setup employed here follows the original setup proposed by~\citet{rabault2019artificial} based on the ``2D-2'' testcase in~\cite{schafer1996benchmark}.
It describes the flow around a two-dimensional cylinder immersed in a channel with height $H=4.1$ and length $L=22$ as depicted in \cref{fig:simulation:setup}.
The cylinder is offset from the channel centerline by $5\%$ of the cylinder diameter $D=1$ to facilitate the onset of vortex shedding.
The Reynolds number is chosen as $\mathrm{Re}_D= D U_b/\nu=100$ with respect to $D$ and the bulk velocity $U_b=1$.
A parabolic velocity profile is imposed at the inflow boundary via a Dirichlet boundary condition and follows the form
\begin{equation}
  u(x=0, y) = U_b \frac{6y\,(H-y)}{H^2}.
\end{equation}
The upper and lower walls of the channel are modeled as adiabatic no-slip walls and the outflow condition is prescribed via a constant pressure outlet as introduced in \cite{carlson2011FUN3D}.

The immersed cylinder surface is modeled as an adiabatic wall, identical to the upper and lower channel walls.
For the controlled case, two synthetic jets with angular width $\omega$ are placed at the top and bottom of the cylinder.
These jets are implemented by imposing a non-zero normal velocity component at the cylinder wall $u_\perp$, while the remaining components of the state vector are computed analogously to an adiabatic no-slip wall.
The velocity profile of each jet follows a cosine distribution that reaches the peak velocity at the jets' center and vanishes at its bounds at $\pm \frac{\omega}{2}$.
This yields the normal velocity component at the cylinder wall as
\begin{equation}
  u_\perp(\varphi) =
  \begin{cases}
    Q_i \frac{\pi}{2\omega D^2}\cos\left(\frac{\pi}{\omega} (\varphi-\varphi_i)\right) & \text{if }|\varphi-\varphi_i|\leq \frac{\omega}{2}\\
    0                                                                                  & \text{otherwise}
  \end{cases},
\end{equation}
where $\varphi$ denotes the radial coordinate on the cylinder surface with respect to its center.
Here, $\varphi_i = \pm \pi/2$ denotes the locations of the two jets with $i=1,2$, and $Q_i$ is the scalar controlling the jet strength of the respective jet.
Throughout this work, we enforce a net-zero mass flux by setting the sum of the jet strengths to zero, which finally yields a single scalar quantity
\begin{equation}
  \hat{Q}=Q_1=-Q_2,
  \label{eq:zero_mass_flux}
\end{equation}
to control both jets.
The reader is referred to \cite{rabault2019artificial} for further details on the test case setup.

The domain is discretized using a high-order, block-structured mesh with 150 quadrilateral, curved elements using a fourth-order polynomial representation of the geometry.
The simulation is then performed using a polynomial order of $N=4$ with a Discontinuous Galerkin~(DG) method on collocated Legendre--Gauss interpolation and integration points~\cite{krais2021flexi}.
This results in a total of \num{3750} degrees of freedom per solution variable.
In order to simulate the incompressible flow with the compressible flow solver FLEXI~\cite{krais2021flexi},  a Mach number has to be imposed to recover the background pressure and thus a fully compressible flow state.
For the present study, the Mach number is set to $\mathrm{Ma}=0.2$.
A validation of the present simulation setup against reference results from the literature with a more in-depth discussion of compressibility effects is given in \cref{sec:validation}.

Common metrics of evaluation are the drag, lift, and total force coefficients, defined as
\begin{equation}
  C_D = \frac{F_x}{U_b\rho/2} \; , \quad C_L = \frac{F_y}{U_b\rho/2} \; , \quad C_F = \frac{\sqrt{F_x^2 + F_y^2}}{U_b\rho/2},
\end{equation}
respectively.
Here, $F_x$ and $F_y$ denote the forces acting on the cylinder in $x$- and $y$-direction, respectively, and the density $\rho=1$.
Similarly, we report all results in the following based on the non-dimensional time $t^*=t U_b/D$.

\subsection{Reinforcement Learning Setup}
\label{subsec:RLsetup}
Unlike other machine learning methods, reinforcement learning seeks to solve a class of sequential decision tasks known as \textit{Markov Decision Processes} (MDP), by interacting with an external environment with no prior information on what a good decision policy might be.
Here, an \textit{agent} is embedded in said environment, with the environment itself residing in a state $\ppvec{s}_t$ at step $t$.
The agent observes the environment through partial \textit{observations} $\ppvec{o}_t=\ppvec{o}_t(\ppvec{s}_t)$ and suggests an \textit{action} $\ppvec{a}_{t}$ based on these observations.
For each action, the environment transitions into a new state $\ppvec{s}_{t+1}$ and the agents receives a scalar \textit{reward} $r_{t+1}$ based on this transition.
The reward is computed using a reward function $r_{t+1}=R(\ppvec{s}_t,\ppvec{a}_t,\ppvec{s}_{t+1})$ based on how the suggested action affected the evolution of the environment's state.

Deep Reinforcement Learning (DRL) refers to the branch of RL where the agent is represented by a deep neural network.
While numerous training strategies for DRL tasks exist~\cite{schulman2017proximal, lillicrap2015continuous, barth2018distributed, gu2016continuous}, we employ the Proximal Policy Optimization (PPO)~\cite{schulman2017proximal} algorithm, for two major reasons.
First, PPO is mathematically and computationally simpler than other DRL methods while being robust and yielding competitive performance for a wide range of problems.
Second, PPO is well established in the general RL literature and for AFC applications, and has standardized and widely tested implementations readily available for use.

The PPO method is episodic, meaning that it waits for the environment to reach a terminal state before beginning to process all the data and perform learning.
This means that it only learns to optimize the policy within these episode durations.
Therefore, the episode length has to be chosen sufficiently long to allow the agent to learn policies that yield stable and favorable results also for longer time scales.
PPO also requires training two networks in tandem, a \textit{policy/actor} network and a \textit{critic/value} network.
Both take as input the observations $\ppvec{o}_t$.
The actor network learns the control policy, and its outputs are the actions that are passed to the environment.
The critic network learns to predict the \textit{discounted sum of future rewards} for a state.
This value is used in the policy loss computation, as a measure to reduce its variance and improve learning stability.
For further details, the reader is referred to standard RL resources such as \cite{schulman2017proximal, sutton2020reinforcement} since these are not the focus of the present work.

For this study, we use the simulation environment as described in \cref{subsec:simulation} as the environment for the agent to interact with.
For this, we define the MDP as follows.

\subsubsection*{Observations $\ppvec{o}_t$}\label{sec:RLsetup_obs}
The full state of the environment $\ppvec{s}_t$ would correspond to the full flow field of the simulation entailing the solution at each grid point.
In this work, the agent obtains only partial information of the flow field via 11 pressure probes that are dispersed in the domain, as shown in \cref{fig:simulation:setup}.
The policy receives the pressure deviations from the reference pressure at these locations as input, i.e. $\Delta p_i = p(\ppvec{x}_{\text{probe},i})- p_{b}$ with $i=1,\ldots,11$.
In addition, the GNN policy also receives the normalized coordinates of the probe position $\ppvec{x}^*_i = (\ppvec{x}_{\text{probe},i}- \ppvec{x}_{\text{cylinder,center}})/D$ as input.
This is a vector with two entries - the first the wall-parallel component and as second entry the wall-normal component.
The reason for this is that the GNN policy requires some form of positional encoding to distinguish between the upper and lower halves of the domain.
This requirement is discussed in further detail in \cref{sec:RLsetup:policynet}.
In contrast, MLP policy networks can learn this positional information from the order of the probes in its input vector.
However, this characteristic is also one of the MLP's major shortcomings since it then learns to rely explicitly on the ordering of the input vector for this information, unable to function accurately when this is permuted.
By formulating the position with respect to the cylinder center and the wall orientation, we ensure the policy remains invariant to the choice of the coordinate system.

\subsubsection*{Actions $\ppvec{a}_t$}
The action is defined identically to the reference work~\cite{rabault2019artificial}.
The agent controls the strength of the two synthetic jets at the top and bottom of the cylinder as described in \cref{subsec:simulation}.
By imposing the net-zero mass flux condition given in \cref{eq:zero_mass_flux}, the agent only needs to control a single scalar quantity $\hat{Q}$. 
To ensure that the agent keeps the actuation within a reasonable range, the action prediction of the neural networks is limited using a sigmoid activation function and scaled linearly to the range $\hat{Q} \in [-0.067 Q_{\text{ref}},0.067 Q_{\text{ref}}]$, where $Q_{\text{ref}}$ is the mean mass flow across the cylinder diameter based on the inflow condition.

\subsubsection*{Reward $r_t$}

In this work, we modify the reward function proposed by~\citet{rabault2019artificial}, which is given by
\begin{equation}
    r_t = - \langle C_D \rangle - \alpha \left|\langle C_L\rangle\right| ,
    \label{eq:old_reward}
\end{equation}
where $\langle \cdot \rangle$ denotes the time-averaged value of the respective quantity and $\alpha$ is a weighting factor that determines how strongly the agent is penalized for introducing a mean lift component.
A major drawback of this reward function is that the drag coefficient $C_D$ is always positive and oscillates only slightly around a constant mean value  $\langle C_D\rangle \approx 2.8$ for the uncontrolled flow case. This introduces a significant bias towards strongly negative rewards giving cumulative returns per episode of approximately $-230$ at the start of the training.
This is undesirable since it requires the critic to learn to predict such low returns during the first several training epochs.
Since the advantage estimates are based on the critic's predictions, this can lead to arbitrary changes and slow convergence of the policy network.

Instead, we propose a modified reward function based on the relative drag reduction compared to the uncontrolled case, similar to the reward function proposed by \citet{cavallazzi2024deep}.
For this, we compute the reward as
\begin{equation}
  r_t = \frac{\langle C_{D} \rangle^{\text{no control}} - \langle C_{D}\rangle }{\langle C_{D} \rangle^{\text{no control}} - C_{D,\text{min}}} - \alpha |\langle C_L\rangle | ,
    \label{eq:reward_mod}
\end{equation}
where, the parameters $\langle C_{D} \rangle^{\text{no control}}$ and $C_{D,\text{min}}$ are the mean drag coefficient of the uncontrolled case and the expected minimum achievable drag coefficient, respectively.
The mean drag and lift, i.e.\ $\langle C_D \rangle$ and $\langle C_L \rangle$, are computed using an exponential moving average during the simulation to give more weight to the more recent values.

\begin{table}[htb]
    \centering
    \begin{tabular}{llp{4cm}}
        \toprule
        \textbf{Hyperparameter}         & \textbf{Value}        & \textbf{Description} \\
        \midrule
        $\mathrm{Ma}$                   & 0.2                   & Flow Mach number. \\
        $\mathrm{Re}$                   & $100$                 & Flow Reynolds number. \\
        \midrule
        $t_{end}$                       & 20                    & Simulation end time. \\
        $\Delta t_{RL}$                 & 0.25                  & Control update time interval. \\
        $C_{D,\text{min}}$              & 2.5                   & Minimum drag coefficient - used for reward scaling. \\
        $\langle C_{D} \rangle^{\text{no control}}$ & 2.8       & Mean drag coefficient of the uncontrolled case - used for reward scaling. \\
        \midrule
        $n_{env}$                       & 32                    & No. of episodes per update. \\
        $\eta^{actor}$                  & \cref{fig:base:learningrate}         & Learning rate for actor. \\
        $n_{epochs}^{actor}$            & 40                    & Max. training epochs for actor per iteration. \\
        $\mathcal{T}_{ES}^{actor}$      & 0.025                 & Early stopping threshold for actor.\\
        $\eta^{critic}$                 & 0.0005                & Learning rate for critic. \\
        $n_{epochs}^{critic}$           & 100                   & Max. training epochs for critic per iteration. \\
        ${n}_{patience}^{critic}$       & 5                     & Early stopping patience value for critic.\\
        \midrule
        $\sigma^{actor}$ & 0.02 & Standard deviation of the action distributions.\\
        \midrule
        Neurons                             & 256                   & No. of neurons in the MLP actor and critic networks. \\
        Layers                              & 2                     & No. of layers in the MLP actor and critic networks. \\
        \bottomrule
    \end{tabular}
    \caption{Hyperparameters and their values.}%
    \label{tab:hyperparameters}
\end{table}

\subsection{Policy Networks}
\label{sec:RLsetup:policynet}

\begin{table}[tb]
    \centering
    \begin{tabular}{lcrr}
        \toprule
        \multirow{2}{*}{\textbf{Layer}}  & \multirow{2}{*}{\textbf{Node-Local}} & \textbf{Learnable}  & \multirow{2}{*}{\textbf{Dimension}} \\
                                         &                                      & \textbf{Parameters} &                                     \\
        \midrule
        Input Layer                      &                                  Yes &                   - &  $(N, 3)$ \\
        Dense Encoder                    &                                  Yes &            \num{64} &  $(N,16)$ \\
        Message-Passing                  &                                   No &          \num{4352} & $(N,256)$ \\
        Message-Passing                  &                                   No &         \num{65792} & $(N,256)$ \\
        Dense Decoder                    &                                  Yes &          \num{4112} &  $(N,16)$ \\
        Graph-Average                    &                                   No &                   - &    $(16)$ \\
        Dense Output                     &                                    - &            \num{17} &     $(1)$ \\
        \midrule
        \textbf{Total}                   &                                    - &         \num{74337} &         - \\
        \bottomrule
    \end{tabular}
    \caption{Architecture of the employed GCNN policy for a graph with $N$ nodes. The number of trainable weights and the output dimensions are given for each layer.}%
    \label{tab:gnn_architecture}
\end{table}

\begin{figure}[tb]
    \centering
    \tikzsetnextfilename{fig_mlp}
    \tikzset{%
    every neuron/.style={
        circle,
        draw,
        minimum size=0.05cm
    },
    neuron missing/.style={
        draw=none,
        scale=1.5,
        text height=0.333cm,
        execute at begin node=\color{black}$\vdots$
    },
}

\begin{tikzpicture}[x=0.8cm, y=0.8cm, >=stealth]

    \foreach \m/\l [count=\y] in {1,2,missing,3}
    \node [every neuron/.try, neuron \m/.try] (input-\m) at (0,2.75-\y) {};

    \foreach \m [count=\y] in {1,2,missing,3,4}
    \node [every neuron/.try, neuron \m/.try ] (hidden-\m) at (2,4-\y*1.25) {};

    \foreach \m [count=\y] in {1,2,missing,3,4}
    \node [every neuron/.try, neuron \m/.try ] (hidden2-\m) at (4,4-\y*1.25) {};

    \foreach \m [count=\y] in {1}
    \node [every neuron/.try, neuron \m/.try ] (output-\m) at (6,1.5-\y) {};

    \foreach \l [count=\i] in {1,2,11}
    \draw [<-] (input-\i) -- ++(-1,0)
    node [above, midway] {$RP_{\l}$};


    \foreach \l [count=\i] in {1}
    \draw [->] (output-\i) -- ++(1,0)
    node [above, midway] {$\hat{Q}$};

    \foreach \i in {1,...,3}
    \foreach \j in {1,...,4}
    \draw [->] (input-\i) -- (hidden-\j);

    \foreach \i in {1,...,4}
    \foreach \j in {1,...,4}
    \draw [->] (hidden-\i) -- (hidden2-\j);

    \foreach \i in {1,...,4}
    \foreach \j in {1}
    \draw [->] (hidden2-\i) -- (output-\j);

    \foreach \l [count=\x from 0] in {Input, Hidden, Hidden, Ouput}
     \node [align=center, above] at (\x*2,3) {\l \\ layer};
\end{tikzpicture}
    \caption{MLP policy network architecture using 2 hidden layers with 256 neurons each.}
    \label{fig:RL:mlp_architecture}
\end{figure}
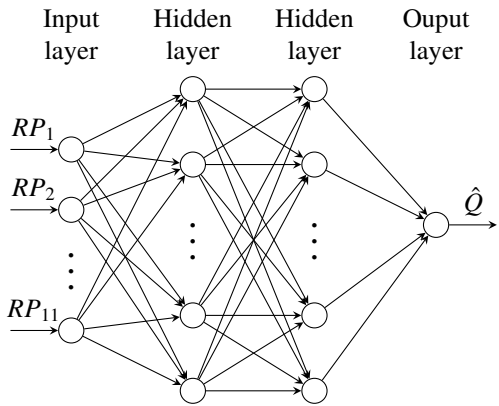

\begin{figure}[tb]
    \centering
    \tikzsetnextfilename{fig_gnn_connectivity}
    \input{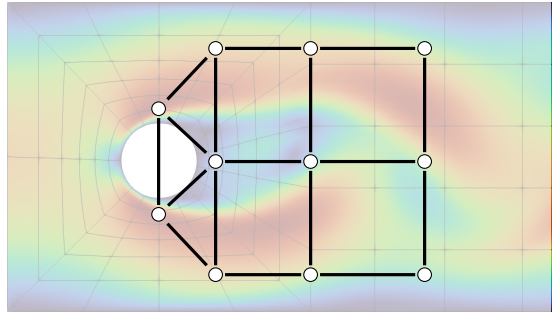}
    \caption{Pressure probes' connectivity used to build the adjacency matrix for the graph net.}
    \label{fig:RL:gnn_connectivity}
\end{figure}

\begin{figure}[tb]
    \centering
    \tikzsetnextfilename{fig_gnn_architecture}
    \input{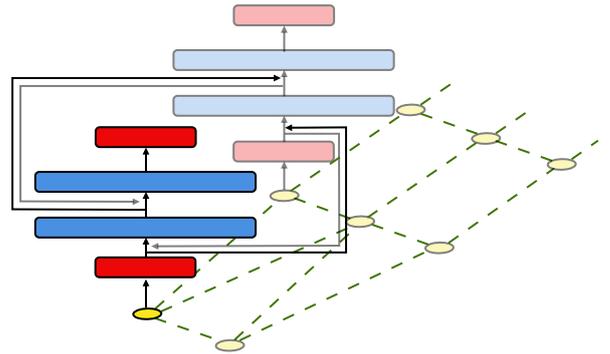}
    \caption{GNN policy network architecture. The data exchange between a pair of neighbors is demonstrated, where the blue boxes represent the message passing layers and the red boxes are standard dense layers. By virtue of their construction, the weights are shared across the nodes, and hence this architecture can be considered to induce the MARL paradigm.}
    \label{fig:RL:gnn_architecture}
\end{figure}

Two policy network architectures are evaluated in this study, a standard multi-layer perceptron (MLP) and a Graph Convolutional Neural Network (GCNN)~\cite{kipf2017semisupervised}.
The policy is assumed to be distributed normally. The output of the network is used as the mean of the distribution and its standard deviation $\sigma^{\text{actor}}$ is chosen to be state-independent, and thus a fixed hyperparameter.
During training, the actions in each step are drawn from this parametrized Gaussian distribution for generating the trajectories. During evaluation however, the action is sampled greedily from the distribution, which corresponds to selecting the most probable value of the distribution.
For a Gaussian distribution, this amounts to selecting the distribution mean.
Greedy evaluation thus results in a deterministic policy.

The MLP consists of an initial input layer comprising the concatenated observations received from the environment, followed by two fully-connected hidden layers and a dense output layer as illustrated in~\cref{fig:RL:mlp_architecture}.
%
The architecture of the GNN is summarized in~\cref{tab:gnn_architecture}.
In a first step, the GNN trunk processes the nodal states to learn an initial latent embedding vector for each node $n_i$ using a fully connected layer that computes
\begin{equation}
  \ppvec{h}^1_{i}= \phi^1 \left( \ppmat{W}^1 \ppvec{o}_i + \ppvec{b}^1 \right),
  \label{eq:graph:encoder}
\end{equation}
where $\ppvec{o}_i$ is the input observation vector for node $n_i$, $\ppmat{W}^1$ and $\ppvec{b}^1$ are the trainable weight matrix and bias vector of the layer, respectively, and $\phi^1$ is the nonlinear activation function.
In our case, this initial embedding has the dimension $\ppvec{h}^1_{i} \in \mathds{R}^{16}$.
This is followed by two message passing layers that update the embedding vector based on embeddings of the neighboring nodes following
\begin{equation}
    \ppvec{h}^{l+1}_i = \phi^l \left[ \left( \ppvec{h}^{l}_i + \sum_{j \in \mathcal{N}_i} c_{ij} \ppvec{h}^{l}_j\right) \ppmat{W}^l + \ppvec{b}^l \right].
    \label{eq:setup:graphconv}
\end{equation}
Here, $\ppvec{h}^{l}_i$ is the input vector at the $l^{th}$ message-passing layer at node $n_i$, $\mathcal{N}_i=\mathcal{N}(n_i)$ is the neighborhood of node $n_i$, and $c_{ij}$ are the weights assigned to the connection between node $n_i$ and a node $n_j \in \mathcal{N}_i$.
Again, the matrix $\ppmat{W}^l$ denotes the trainable weight matrix and $\ppvec{b}^l$ the bias vector of this layer, with $\phi^l$ as its nonlinear activation function.
A GCNN computes the weights $c_{ij}$ and encodes the neighborhood information using the adjacency matrix $\ppmat{A}\in\mathds{R}^{NxN}$ of a graph comprising $N$ nodes as detailed by \citet{kipf2017semisupervised}.
The adjacency matrix is defined as
\begin{equation}
  A_{ij} =
  \begin{cases}
    1 & \text{if } n_j \text{ is a direct neighbor of } n_i \\
    0 & \text{otherwise}.
  \end{cases}
\end{equation}
and thus encodes the connectivity and structure of the graph.
For undirected graphs as in our case, the adjacency matrix is symmetric.
We construct $\ppmat{A}$ for our application to the cylinder case using the connectivity displayed in \cref{fig:RL:gnn_connectivity}.
At the end of the message-passing layers, similar to the encoding layer \cref{eq:graph:encoder}, a dense layer transforms each nodal state into an intermediate-size (16 neurons) representation, which is then averaged across the graph. A final dense layer then maps the graph-averaged state to the final output size.

Owing to this averaging operation, the GNN is unable to distinguish between flow phenomena in the lower half of its domain and those in the upper half (as noted in \cref{sec:RLsetup_obs}). This is especially crucial in our case where the control action itself is directed, and depends on the policy network being able to distinguish between such cases. The MLP does not suffer from this problem because it learns implicit positional information based on the ordering of its input which is also one of its major shortcomings. To remedy this, the flow-parallel cylinder-centered normalized coordinates of the pressure probes $\ppvec{x}^*_i$, in addition to the pressure values, are provided to the GNN as inputs.

\subsection{Critic Networks}

Of interest to the present study is the architecture of the critic network. A simple MLP is employed, mirroring the architecture of the MLP policy network, with the concatenated observations as its input, followed by 2 dense layers with 256 neurons and a final dense layer leading to the output of the network. This output is passed through a hyperbolic tangent activation to scale it to $[-1, 1]$ (the predicted future rewards can be negative), and then scaled by a single learnable weight. The same critic network architecture is employed in the training of both the MLP and GNN policy networks, to avoid giving either an unfair advantage (or disadvantage).

\subsection{Implementation Details}
The machine learning part of this work is based on Relexi~\cite{kurz2022relexi} which is an open-source reinforcement learning framework designed for HPC environments.
It is written in Python and based on TensorFlow's \cite{tensorflow2015-whitepaper} RL library, TF-Agents~\cite{TFAgents}.
Relexi enables the use of RL for computationally intensive simulations (such as CFD), by coupling HPC simulation codes with the TF-Agents library via the SmartSim package~\cite{partee2022using} for fast HPC-scalable environments.
Relexi automates the workload distribution and managment of parallelized simulation instances across allocated HPC resources.
The flow solver FLEXI~\cite{krais2021flexi} is used to run the simulations. It is a high-order accurate simulation framework designed to solve partial differential equations with a particular emphasis on computational fluid dynamics. It employs the discontinuous Galerkin spectral element method (DGSEM), which facilitates high-order accuracy and supports fully unstructured hexahedral meshes. Developed by the Numerics Research Group (NRG) at the University of Stuttgart’s Institute of Aerodynamics and Gasdynamics, FLEXI has demonstrated exceptional scalability - efficiently operating on large-scale applications on over \num{500000} computing cores \cite{blind2022performance,blind2024wall,durrwachter2021efficient} and recently has also been adapted for GPU systems~\cite{kurz2025galaexi}.


\begin{figure}[tb]
    \centering
    \tikzsetnextfilename{fig_learningrate_MLP}
    \begin{tikzpicture}[font=\footnotesize]
  \begin{groupplot}[
        cycle list/Dark2,
        group style={
          group size=2 by 1,
          horizontal sep={0.06\textwidth},
          vertical sep={0.06\textwidth}
        },
        width=0.99\linewidth,
    ]

  \def\constiter{100} 
  \def\maxiter{140}
  \def\figwidth{0.98\linewidth}
  \def\figheight{0.5\linewidth}
  \def\nthpoint{1} 
  \def\epochs{1}

  \nextgroupplot[
    grid=both,
    grid style={line width=.1pt, draw=gray!5},
    major grid style={line width=.2pt,draw=gray!25},
    xmin = 0, xmax =\maxiter,
    ymode=log,
    width=\figwidth,
    height=\figheight,
    xlabel={Iterations},
    ylabel={$\eta$},
    legend pos=north east,
    legend cell align={left},
  ]

  \addplot+[thick] table[x expr=\thisrowno{1}/\epochs,x index={1},y index={2},col sep=comma, each nth point=\nthpoint, filter discard warning=false]{./tikz/results_base/MLP_Plot_data-_flash_2025_02_24_relexi_run_001/run-FlowControl_2D_Cylinder_train-tag-LearningRate_learning_rate_policynet.csv};
  \addlegendentry{Actor learning rate ($\eta^{actor}$)}
    \draw[-latex] (axis cs:\constiter+5,2e-4) -- (axis cs:\maxiter-5,2e-4) node[midway,above,anchor=south] {\footnotesize $\eta^{\text{actor}}=10^{-4}$};
    \draw[dashed] (axis cs:\constiter,1e-5) -- (axis cs:\constiter,1e-2);

  \end{groupplot}
\end{tikzpicture}
    \caption{Actor's learning rate decay throughout training. The learning rate becomes constant after 100 iterations with the last value (i.e. $10^{-4}$ in this case).}%
    \label{fig:base:learningrate}
\end{figure}
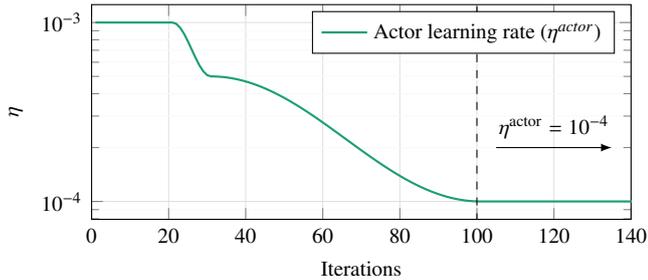

Note that unlike other studies, our sampled environments start from different points on the vortex shedding cycle, thus making the learned policy more robust to starting conditions, and providing ample variety of state-action-reward tuples for training.
We provide a set of 8 start files for the environments to choose from at the beginning of each trajectory-sampling iteration.

Further, our PPO implementation implements Kullback-Leibler (KL) divergence based early stopping for the policy networks every iteration, which is standard practice in RL research~\cite{schulman2017proximal, dossa2021empirical, raffin2021stable, sun2022you}.
Essentially, if the KL divergence between the distributions from the beginning of the update iteration and the end of an update epoch exceeds a threshold $\mathcal{T}_{ES}^{actor}$, actor-net training is halted until trajectories are resampled.
This helps to limit the policy from venturing too far from the original (i.e. the one used to generate the trajectories) and avoids harming returns during updates, since clipping is known to be insufficient in this regard.
Inspired by this and standard DL early stopping, we also implement simple patience-based early stopping for the critic network training, where if the value-function loss does not decrease for $n_{patience}^{critic}$ epochs, critic-net training is halted until trajectories are resampled.
Both these additions help with speeding up training (as unnecessary training epochs are avoided) and preserving performance.
Lastly, we use a decaying learning rate (see \cref{fig:base:learningrate}) for the actor networks to modulate performance, as noted in \citet{engstrom2019implementation}.
The specific values for all the relevant hyperparameters used in this study are summarized in \cref{tab:hyperparameters}.

\section{Results}
\label{sec:results}

The following section presents the main findings of this study and is structured as follows.
First, the training behavior of the MLP and GNN policies are discussed in \cref{sec:results:training} and validated against the reference study by~\citet{rabault2019artificial} in \cref{sec:results:validation}.
Subsequently, the performance of the trained policies is analyzed in detail in \cref{sec:results:performance} and the permutation invariance of the GNN policy is verified empirically in \cref{sec:results:permutation}.
Finally, we discuss the control strategies learned by the policy networks in \cref{sec:results:strategy}.

\subsection{Training Behavior}%
\label{sec:results:training}

\begin{figure}
    \centering
    \tikzsetnextfilename{fig_training_reward_cylinder_base}
    \begin{tikzpicture}[font=\footnotesize]
  \begin{groupplot}[
        cycle list/Dark2,
        group style={
          group size=1 by 2,
          horizontal sep={0.06\textwidth},
          vertical sep={0.02\textwidth}
        },
        width=0.99\linewidth,
        grid=both,
        grid style={line width=.1pt, draw=gray!5},
        major grid style={line width=.2pt,draw=gray!25},
        xmin = 0, xmax =\maxiter,
        width=\figwidth,
        height=\figheight,
    ]

  \def\maxiter{2000}
  \def\figwidth{0.94\linewidth}
  \def\figheight{0.5\linewidth}
  \def\nthpoint{1} 
  \def\epochs{1}

  \nextgroupplot[
    xticklabels={,,},
    ylabel={Reward - MLP},
    legend pos=south east,
    legend cell align={left},
    legend style={font=\scriptsize},
  ]

  \addplot+[forget plot,name path=maxreturn, draw=none] table[x expr=\thisrowno{1}/\epochs,x index={1},y index={2},col sep=comma, each nth point=\nthpoint, filter discard warning=false]{./tikz/results_base/MLP_Plot_data-flash_2025_02_27-run00-relexi_run_004/run-FlowControl_2D_Cylinder_train-tag-MetricsTrain_max_return.csv};
  \addplot+[forget plot,name path=minreturn, draw=none] table[x expr=\thisrowno{1}/\epochs,x index={1},y index={2},col sep=comma, each nth point=\nthpoint, filter discard warning=false]{./tikz/results_base/MLP_Plot_data-flash_2025_02_27-run00-relexi_run_004/run-FlowControl_2D_Cylinder_train-tag-MetricsTrain_min_return.csv};
  \addplot+[forget plot,opacity=0.3] fill between [of = maxreturn and minreturn];
  \addplot+[thick] table[x expr=\thisrowno{1}/\epochs,x index={1},y index={2},col sep=comma, each nth point=\nthpoint, filter discard warning=false]{./tikz/results_base/MLP_Plot_data-flash_2025_02_27-run00-relexi_run_004/run-FlowControl_2D_Cylinder_train-tag-MetricsTrain_avg_return.csv};
  \addlegendentry{Training}
  \addplot+[thick] table[x expr=\thisrowno{1}/\epochs,x index={1},y index={2},col sep=comma, each nth point=\nthpoint, filter discard warning=false]{./tikz/results_base/MLP_Plot_data-flash_2025_02_27-run00-relexi_run_004/run-FlowControl_2D_Cylinder_train-tag-MetricsEval_eval_avg_return.csv};
  \addlegendentry{Eval}

  \nextgroupplot[
    xlabel={Iterations},
    ylabel={Reward - GNN},
  ]
  \addplot+[forget plot,name path=maxreturn, draw=none] table[x expr=\thisrowno{1}/\epochs,x index={1},y index={2},col sep=comma, each nth point=\nthpoint, filter discard warning=false]{./tikz/results_base/GNN_Plot_data-HAWK_2025_02_17-run03-relexi_run_014/run-FlowControl_2D_Cylinder_train-tag-MetricsTrain_max_return.csv};
  \addplot+[forget plot,name path=minreturn, draw=none] table[x expr=\thisrowno{1}/\epochs,x index={1},y index={2},col sep=comma, each nth point=\nthpoint, filter discard warning=false]{./tikz/results_base/GNN_Plot_data-HAWK_2025_02_17-run03-relexi_run_014/run-FlowControl_2D_Cylinder_train-tag-MetricsTrain_min_return.csv};
  \addplot+[forget plot,opacity=0.3] fill between [of = maxreturn and minreturn];
  \addplot+[thick] table[x expr=\thisrowno{1}/\epochs,x index={1},y index={2},col sep=comma, each nth point=\nthpoint, filter discard warning=false]{./tikz/results_base/GNN_Plot_data-HAWK_2025_02_17-run03-relexi_run_014/run-FlowControl_2D_Cylinder_train-tag-MetricsTrain_avg_return.csv};
  \addplot+[thick] table[x expr=\thisrowno{1}/\epochs,x index={1},y index={2},col sep=comma, each nth point=\nthpoint, filter discard warning=false]{./tikz/results_base/GNN_Plot_data-HAWK_2025_02_17-run03-relexi_run_014/run-FlowControl_2D_Cylinder_train-tag-MetricsEval_eval_avg_return.csv};

  \end{groupplot}
\end{tikzpicture}
    \caption{Evolution of the training and evaluation return during the training for the MLP (\textit{top}) and the GNN (\textit{bottom}) models. As can be observed, the training is robust and converges smoothly to a stable maximum for both cases.}%
    \label{fig:res:reward}
\end{figure}
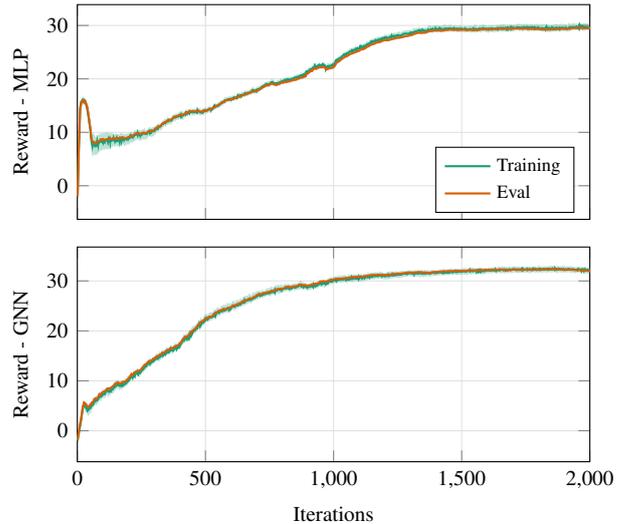

To investigate the training behavior of the GNN and MLP models, the evolution of the cumulative rewards during the training is shown in \cref{fig:res:reward}.
At the beginning of the training, both models still exhibit random initial weights and collect very similar, slightly negative rewards.
Starting after this initial state, the training process appears to be very stable for both models, with the cumulative rewards evolving monotonically and converging to a stable maximum after approximately 1500 iterations.
Moreover, the variation of the collected rewards across the parallel training environments appears to be negligible in comparison to the overall improvement.
This robustness is also reflected by the fact that the greedy evaluation of the policy yields comparable performance to the stochastic training environments, which shows that no overfitting or mismatch between the stochastic training and the greedy evaluation runs occur.
Overall, the GNN is observed to learn faster and to reach a higher final return compared to the MLP.
We attribute this to the invariant nature of the GNN and its weight-sharing mechanism, which appears to improve the training efficiency per training data sample.
Before investigating the behavior of the trained policies in detail, we first validate our overall setup by comparing our results to the reference study by~\citet{rabault2019artificial} in the following.

\subsection{Validation with Reference}%
\label{sec:results:validation}

First, we validate the performance of the trained MLP policy against the results reported in the original study~\cite{rabault2019artificial}.
To account for the different baseline drag coefficients due to compressibility effects (cf.~\cref{sec:validation}), we compare the drag reduction as follows.
Besides the baseline no-control case, we compute a second reference case in which we impose a symmetry condition in the center cylinder plane to suppress the vortex shedding as detailed in \cref{sec:symmetric} and also reported in the reference study. To distinguish it from the baseline no-control case, this case is referred to as the `symmetric' case for the remainder of this study.
Based on this, we compute an \textit{improvement factor} $r_{X}$ for a coefficient $C_X$ as
\begin{equation}
  r_{X} = \frac{
    \langle{C_X\rangle}^{\text{no control}} - \langle{C_X}\rangle^{\text{AFC}}
    }{
    \langle{C_X\rangle}^{\text{no control}} - \langle{C_X}\rangle^{\text{sym}}
    } -1,
    \label{eq:res:reduction_factor}
\end{equation}
where $\langle{C_X\rangle}^{\text{no control}}$, $\langle{C_X}\rangle^{\text{sym}}$ and $\langle{C_X}\rangle^{\text{AFC}}$ are the time-averaged coefficients of the baseline case without control, the symmetric case without any vortex shedding and the case with AFC, respectively.
Here, $r_{X}=0$ is recovered if the AFC case achieves the same reduction as the symmetric case and $r_{X}>0$ denotes that the AFC case is able to decrease the coefficient below the one reported for the symmetric case.
To elucidate, the improvement factor compares the suppression of $C_X$ through AFC w.r.t. the symmetric case without any vortex shedding - which can be seen as a good approximation to the maximum drag reduction achievable through AFC.

The time-averaged $\langle{C_D}\rangle$ for our MLP policy was computed by running a simulation with the trained control policy up to $t^*=100$ and averaging over the interval $t^*\in[50,100]$ to capture the long-term effects and to omit any influence of the initial transient.
The resulting time-averaged drag coefficients of our MLP policy, the baseline case and the symmetric case, and those from the reference study \cite{rabault2019artificial} are summarized in~\cref{tab:res:validation}.
Based on these values, we also compute the improvement factors $r_D$ for each of the two studies.

Our MLP policy achieves an improvement in drag reduction of $r_{D}=9.09\%$ over the symmetric case, which indicates that the learned policy not only suppresses the vortex shedding but finds a more intricate control strategy to even undercut the results of the symmetric case.
The reference study also reports a significant reduction in drag by the MLP control law in comparison to the no-control case by reducing the drag coefficient from $\langle{C_D}\rangle=3.20$ to $\langle{C_D}\rangle=2.99$.
However, the MLP there is not able to reduce the drag down to the level of the symmetric case, which is $\langle{C_D}\rangle=2.93$, resulting in a $r_D=-22.21\%$.

The MLP trained in the present work thus not only recreates, but surpasses the performance of the equivalent 11-probe case of the original reference.
Even more remarkably, this policy is able to achieve this reduction using significantly fewer model parameters, with the reference study employing roughly \num{300000} learnable parameters while the present work uses only around \num{75000}. Although this direct comparison is not the goal of this study, it nonetheless serves as a validation of our methodology and results. We attribute these improvements to the improved training procedure with the KL-divergence early stopping and the gradually decaying learning rate.
However, the number of iterations required to reach convergence is noticeably higher than in the original study.
This is likely due to their reuse of experience via a replay buffer, which is not used in this study.
Based on these results, we now investigate the performance of the trained MLP and GNN policies in more detail.


\begin{table}
    \addtolength{\tabcolsep}{-3pt}
    \begin{tabular*}{\linewidth}{@{\extracolsep{\fill}} lrr}
        \toprule%
                   & \textbf{Present Work}        & \textbf{Reference~\cite{rabault2019artificial}} \\
        \midrule%
        No-Control & $\langle{C_D\rangle} = 2.78$ & $\langle{C_D\rangle} = 3.20$ \\
        Symmetric  & $\langle{C_D\rangle} = 2.67$ & $\langle{C_D\rangle} = 2.93$ \\
        MLP        & $\langle{C_D\rangle} = 2.66$ & $\langle{C_D\rangle} = 2.99$ \\
        \midrule%
        $r_D$  & $\mathbf{+9.09 \%}$         & $-22.21 \%$                    \\
        \bottomrule%
    \end{tabular*}
    \addtolength{\tabcolsep}{3pt}
      \caption{Average drag coefficients $\langle{C_D\rangle}$ computed in the interval $t^*\in[50,100]$ compared to the results of the 11 probe case reported by~\citet{rabault2019artificial}.}%
    \label{tab:res:validation}
\end{table}

\subsection{Performance of the Trained Policies}%
\label{sec:results:performance}

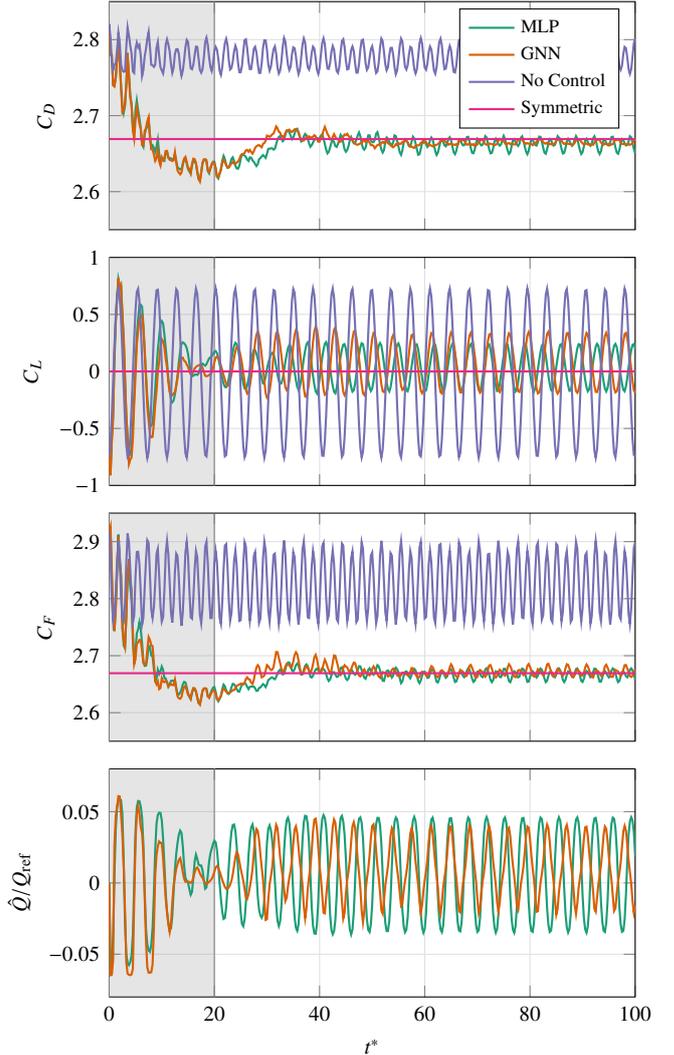
\begin{figure}[htb]
    \centering
    \tikzsetnextfilename{fig_MLP_GNN_NoControl-extended_time}
    \begin{tikzpicture}[font=\footnotesize]
  \begin{groupplot}[
        cycle list/Dark2,
        group style={
          group size=1 by 4,
          horizontal sep={0.06\textwidth},
          vertical sep={0.02\textwidth}
        },
        width=0.99\linewidth,
        grid=both,
        grid style={line width=.1pt, draw=gray!5},
        major grid style={line width=.2pt,draw=gray!25},
        xmin = 0, xmax =\maxiter,
        width=\figwidth,
        height=\figheight,
    ]

  \def\xmaxshade{20}
  \def\xminshade{0}
  \def\maxiter{100}
  \def\figwidth{0.96\linewidth}
  \def\figheight{0.52\linewidth}
  \def\nthpoint{1} 
  \def\epochs{1}

  \nextgroupplot[
    ymin = 2.55, ymax = 2.85,
    xticklabels={,,},
    ylabel={$C_D$},
    legend pos=north east,
    legend cell align={left},
    legend style={font=\scriptsize},
  ]

  \addplot[forget plot,name path=trainxmin,gray] (\xminshade,0) -- (\xminshade,4);
  \addplot[forget plot,name path=trainxmax,gray] (\xmaxshade,0) -- (\xmaxshade,4);
  \addplot[gray,forget plot,opacity=0.2] fill between [of = trainxmin and trainxmax];

  \addplot+[thick] table[x expr=\thisrowno{1}/\epochs,x index={1},y index={2},col sep=comma, each nth point=\nthpoint, filter discard warning=false]{./tikz/results_base/MLP_Plot_data-flash_2025_02_27-run00-relexi_run_004/data.csv};
  \addlegendentry{MLP}

  \addplot+[thick] table[x expr=\thisrowno{1}/\epochs,x index={1},y index={2},col sep=comma, each nth point=\nthpoint, filter discard warning=false]{./tikz/results_base/GNN_Plot_data-HAWK_2025_02_17-run03-relexi_run_014/data.csv};
  \addlegendentry{GNN}

  \addplot+[thick] table[x expr=\thisrowno{1}/\epochs,x index={1},y index={2},col sep=comma, each nth point=\nthpoint, filter discard warning=false]{./tikz/results_base//data_flexi.csv};
  \addlegendentry{No Control}

  \addplot+[thick] table[x expr=\thisrowno{1}/\epochs,x index={1},y index={2},col sep=comma, each nth point=\nthpoint, filter discard warning=false]{./tikz/results_base//data_flexi_baseline.csv};
  \addlegendentry{Symmetric}

  \nextgroupplot[
    ymin = -1.00, ymax = 1.00,
    xticklabels={,,},
    ylabel={$C_L$},
  ]

  \addplot+[forget plot,name path=trainxmin,gray] (\xminshade,-2) -- (\xminshade,2);
  \addplot+[forget plot,name path=trainxmax,gray] (\xmaxshade,-2) -- (\xmaxshade,2);
  \addplot+[gray,forget plot,opacity=0.2] fill between [of = trainxmin and trainxmax];

  \addplot+[thick] table[x expr=\thisrowno{1}/\epochs,x index={1},y index={3},col sep=comma, each nth point=\nthpoint, filter discard warning=false]{./tikz/results_base/MLP_Plot_data-flash_2025_02_27-run00-relexi_run_004/data.csv};

  \addplot+[thick] table[x expr=\thisrowno{1}/\epochs,x index={1},y index={3},col sep=comma, each nth point=\nthpoint, filter discard warning=false]{./tikz/results_base/GNN_Plot_data-HAWK_2025_02_17-run03-relexi_run_014/data.csv};

  \addplot+[thick] table[x expr=\thisrowno{1}/\epochs,x index={1},y index={3},col sep=comma, each nth point=\nthpoint, filter discard warning=false]{./tikz/results_base//data_flexi.csv};

  \addplot+[thick] table[x expr=\thisrowno{1}/\epochs,x index={1},y index={3},col sep=comma, each nth point=\nthpoint, filter discard warning=false]{./tikz/results_base//data_flexi_baseline.csv};

 \nextgroupplot[
    ymin = 2.55, ymax = 2.95,
    xticklabels={,,},
    ylabel={$C_F$},
  ]

  \addplot+[forget plot,name path=trainxmin,gray] (\xminshade,0) -- (\xminshade,4);
  \addplot+[forget plot,name path=trainxmax,gray] (\xmaxshade,0) -- (\xmaxshade,4);
  \addplot+[gray,forget plot,opacity=0.2] fill between [of = trainxmin and trainxmax];

  \addplot+[thick] table[x expr=\thisrowno{1}/\epochs,x index={1},y index={5},col sep=comma, each nth point=\nthpoint, filter discard warning=false]{./tikz/results_base/MLP_Plot_data-flash_2025_02_27-run00-relexi_run_004/data.csv};

  \addplot+[thick] table[x expr=\thisrowno{1}/\epochs,x index={1},y index={5},col sep=comma, each nth point=\nthpoint, filter discard warning=false]{./tikz/results_base/GNN_Plot_data-HAWK_2025_02_17-run03-relexi_run_014/data.csv};

  \addplot+[thick] table[x expr=\thisrowno{1}/\epochs,x index={1},y index={4},col sep=comma, each nth point=\nthpoint, filter discard warning=false]{./tikz/results_base//data_flexi.csv};
  \addplot+[thick] table[x expr=\thisrowno{1}/\epochs,x index={1},y index={4},col sep=comma, each nth point=\nthpoint, filter discard warning=false]{./tikz/results_base//data_flexi_baseline.csv};

  \nextgroupplot[
    ymin = -0.08, ymax = 0.08,
    xlabel={$t^*$},
    ylabel={$\hat{Q} / Q_{\text{ref}}$},
    scaled y ticks=false, 
    yticklabel style={/pgf/number format/fixed, /pgf/number format/precision=2}, 
  ]

  \addplot+[forget plot,name path=trainxmin,gray] (\xminshade,-1) -- (\xminshade,1);
  \addplot+[forget plot,name path=trainxmax,gray] (\xmaxshade,-1) -- (\xmaxshade,1);
  \addplot+[gray,forget plot,opacity=0.2] fill between [of = trainxmin and trainxmax];

  \addplot+[thick] table[x expr=\thisrowno{1}/\epochs,x index={1}, y expr=\thisrowno{4}, y index={4},col sep=comma, each nth point=\nthpoint, filter discard warning=false]{./tikz/results_base/MLP_Plot_data-flash_2025_02_27-run00-relexi_run_004/data.csv};

  \addplot+[thick] table[x expr=\thisrowno{1}/\epochs,x index={1}, y expr=\thisrowno{4}, y index={4},col sep=comma, each nth point=\nthpoint, filter discard warning=false]{./tikz/results_base/GNN_Plot_data-HAWK_2025_02_17-run03-relexi_run_014/data.csv};

  \end{groupplot}
\end{tikzpicture}
    \caption{Long-term evolution of the lift ($C_L$), drag ($C_D$) and total force ($C_F$) coefficients, and the normalized control outputs ($\hat{Q} / Q_{\text{ref}}$) of the agent. The time interval up to $t^*=20$ used for training is shaded. The no-control and the symmetric cases are shown as baselines.}%
    \label{fig:res:policynetoutputs_extended_time}
\end{figure}

In a next step, we investigate the performance of the trained MLP and GNN policies in more detail.
For this, \cref{fig:res:policynetoutputs_extended_time} shows the temporal evolution of the force coefficients $C_L$, $C_D$, and $C_F$ as well as the control output of the policy networks.
The statistics are also computed and summarized in~\cref{tab:res:cl_cd_latterhalf} for a time period $t^*\in[10,20]$ that lies within the training interval.
\Cref{tab:res:cl_cd_latterhalf_extended_time} shows the statistics computed for an extended simulation time $t^*\in[50,100]$ that allows for a more detailed analysis of the long-term behavior of the policy networks.

Most notably, both the MLP and the GNN policy networks achieve qualitatively and quantitatively similar results.
Both models are able to reduce the drag coefficient especially well during the time period up to $t^*=20$, which is the simulation duration used for training.
Similar to the results reported in \cite{rabault2019artificial}, the drag increases again for longer simulation times and reaches a quasi-steady state at around $t^*\approx 40$ for both models.
This robustness of the trained policy is noteworthy as the quasi-steady state differs from the transient behavior observed in the training phase.
This emphasizes the effectiveness of DRL in finding good decision strategies.

Besides the mean values, we also report the amplitudes of the oscillations in the lift and drag coefficients denoted by $\langle{C_X}^{\text{amp}}\rangle$.
For this, we compute the $L^1$-norm with respect to the mean value $\langle{C_X}\rangle$ from the oscillating curves, identify the local peaks and average their resulting values.
While both networks achieve the same total reduction in the average drag, the GNN suppresses the oscillations in the drag coefficient significantly more than the MLP.
Overall, the GNN also exhibits a lower vortex shedding frequency than the MLP agent.
Simultanously, the GNN tends to predict slightly lower control values, which means it has to invest less control effort to achieve the same drag reduction as the MLP policy.
This general trend shifts for the lift force, where the GNN appears to have both a higher mean lift force as well as more pronounced oscillations.
Interestingly, the total force acting on the cylinder $C_F$ for both agents matches the symmetric baseline case without vortex shedding nearly perfectly.
While both networks are able to reduce the drag below the symmetric case, i.e. they yield $r_D>0$, this is at the cost of a minor contribution to the mean lift force that matches exactly the absolute drag reduction.
This implies that the agent learns to balance the terms resembling the drag reduction and lift penalty in the reward function \cref{eq:reward_mod}.

This confirms the initial assumptions that GNNs are suitable network architectures for AFC applicaitons. In a next step, we verify the expected invariance properties of the GNN policy, and the lack thereof in the MLP network.

\begin{table}[t]
    \addtolength{\tabcolsep}{-3pt}
    \begin{tabular*}{\linewidth}{@{\extracolsep{\fill}} lllll}
       \toprule%
                                             & \textbf{MLP}          & \textbf{GNN}          & \textbf{No Control}   & \textbf{Symmetric}     \\
       \midrule%
       $\langle{C_D}\rangle$                 & $2.64$                & $2.64$                & $2.78$                & $2.67$
       \\
       $\langle{C_D}^{\text{amp}}\rangle$    & $1.51 \times 10^{-2}$ & $1.38 \times 10^{-2}$ & $2.45 \times 10^{-2}$ & $0.0$
       \\
       $r_D$                                 & $+27.3 \%$            & $+27.3 \%$            & $--$                  & $--$                 \\
       \midrule
       $\langle{C_L}\rangle$                 & $5.91 \times 10^{-2}$ & $7.27 \times 10^{-3}$ & $0.00$                & $0.00$                \\
       $\langle{C_L}^{\text{amp}}\rangle$    & $2.52 \times 10^{-1}$ & $1.35 \times 10^{-1}$ & $7.39 \times 10^{-1}$ & $0.00$                \\
       \midrule
       $\langle{C_{F}}\rangle$               & $2.64$                & $2.64$                & $2.83$                & $2.67$                \\
       $\langle{C_{F}}^{\text{amp}}\rangle$  & $2.17 \times 10^{-2}$ & $1.74 \times 10^{-2}$ & $7.39 \times 10^{-2}$ & $0.0$ \\
       $r_F$                                 & $+18.8 \%$            & $+18.8 \%$            & $--$                  & $--$                 \\
       \bottomrule%
    \end{tabular*}
    \addtolength{\tabcolsep}{3pt}
      \caption{Time-averaged force coefficients $\langle{C_X}\rangle$ as well as their amplitudes $\langle{C_X}^{\text{amp}}\rangle$, computed and averaged over a time period $t^*\in[10,20]$, which lies within the training time.}%
    \label{tab:res:cl_cd_latterhalf}
\end{table}

\begin{table}[t]
    \addtolength{\tabcolsep}{-3pt}
    \begin{tabular*}{\linewidth}{@{\extracolsep{\fill}} lllll}
        \toprule%
                                             & \textbf{MLP}          & \textbf{GNN}          & \textbf{No Control}   & \textbf{Symmetric}     \\
        \midrule%
        $\langle{C_D}\rangle$                & $2.66$                & $2.66$                & $2.78$                & $2.67$                \\
        $\langle{C_D}^{\text{amp}}\rangle$   & $1.31 \times 10^{-2}$ & $5.02 \times 10^{-3}$ & $2.45 \times 10^{-2}$ & $0.0$ \\
        $r_D$                                & $+9.09 \%$            & $+9.09 \%$            & $--$                  & $--$                 \\
        \midrule
        $\langle{C_L}\rangle$                & $3.80 \times 10^{-2}$ & $7.20 \times 10^{-2}$ & $0.00$                & $0.00$                \\
        $\langle{C_L}^{\text{amp}}\rangle$   & $2.07 \times 10^{-1}$ & $2.59 \times 10^{-1}$ & $7.39 \times 10^{-1}$ & $0.00$                \\
        \midrule
        $\langle{C_{F}}\rangle$              & $2.67$                & $2.67$                & $2.83$                & $2.67$                \\
        $\langle{C_{F}}^{\text{amp}}\rangle$ & $1.36 \times 10^{-2}$ & $1.19 \times 10^{-2}$ & $7.39 \times 10^{-2}$ & $0.0$ \\
        $r_F$                                & $0.00 \%$             & $0.00 \%$             & $--$                  & $--$                 \\
        \bottomrule%
    \end{tabular*}
    \addtolength{\tabcolsep}{3pt}
    \caption{Time-averaged force coefficients $\langle{C_X}\rangle$ as well as their amplitudes $\langle{C_X}^{\text{amp}}\rangle$, computed and averaged over a time period $t^*\in[50,100]$ well after the training time in the quasi-steady limit of the controlled flow.}%
    \label{tab:res:cl_cd_latterhalf_extended_time}
\end{table}


\subsection{Verification of the Permutation Invariance}
\label{sec:results:permutation}

\begin{figure}[h]
    \centering
    \tikzsetnextfilename{fig_ShuffledInput_MLP_GNN}
    \begin{tikzpicture}[font=\footnotesize]
  \begin{groupplot}[
        cycle list/Dark2,
        group style={
          group size=2 by 3,
          horizontal sep={0.03\textwidth},
          vertical sep={0.015\textwidth}
        },
        width=0.99\linewidth,
        grid=both,
        grid style={line width=.1pt, draw=gray!5},
        major grid style={line width=.2pt,draw=gray!25},
        xmin = 0, xmax =\maxiter,
        width=\figwidth,
        height=\figheight,
    ]

  \def\maxiter{20}
  \def\figwidth{0.55\linewidth}
  \def\figheight{0.49\linewidth}
  \def\nthpoint{1} 
  \def\epochs{1}


  \nextgroupplot[
    ymin = 2.55, ymax = 2.85,
    xticklabels={,,},
    ylabel={$C_D$},
    title={MLP},
  ]
  \addplot+[thick] table[x expr=\thisrowno{1}/\epochs,x index={1},y index={2},col sep=comma, each nth point=\nthpoint, filter discard warning=false]{./tikz/results_base/MLP_Plot_data-flash_2025_02_27-run00-relexi_run_004/data.csv};
  \addplot+[thick,dashed] table[x expr=\thisrowno{1}/\epochs,x index={1},y index={2},col sep=comma, each nth point=\nthpoint, filter discard warning=false]{./tikz/results_base/MLP_Plot_data-flash_2025_02_27-run00-relexi_run_004/data_shuffled.csv};

  \nextgroupplot[
    ymin = 2.55, ymax = 2.85,
    xticklabels={,,},
    yticklabels={,,},
    title={GNN},
    legend pos=north east,
    legend cell align={left},
    legend style={font=\scriptsize},
  ]
  \addplot+[thick] table[x expr=\thisrowno{1}/\epochs,x index={1},y index={2},col sep=comma, each nth point=\nthpoint, filter discard warning=false]{./tikz/results_base/GNN_Plot_data-HAWK_2025_02_17-run03-relexi_run_014/data.csv};
  \addlegendentry{Original}
  \addplot+[thick,dashed] table[x expr=\thisrowno{1}/\epochs,x index={1},y index={2},col sep=comma, each nth point=\nthpoint, filter discard warning=false]{./tikz/results_base/GNN_Plot_data-HAWK_2025_02_17-run03-relexi_run_014/data_shuffled.csv};
  \addlegendentry{Shuffled}


  \nextgroupplot[
    ymin = -1.4, ymax = 1.4,
    xticklabels={,,},
    ylabel={$C_L$},
  ]
  \addplot+[thick] table[x expr=\thisrowno{1}/\epochs,x index={1},y index={3},col sep=comma, each nth point=\nthpoint, filter discard warning=false]{./tikz/results_base/MLP_Plot_data-flash_2025_02_27-run00-relexi_run_004/data.csv};
  \addplot+[thick,dashed] table[x expr=\thisrowno{1}/\epochs,x index={1},y index={3},col sep=comma, each nth point=\nthpoint, filter discard warning=false]{./tikz/results_base/MLP_Plot_data-flash_2025_02_27-run00-relexi_run_004/data_shuffled.csv};

  \nextgroupplot[
    ymin = -1.4, ymax = 1.4,
    xticklabels={,,},
    yticklabels={,,},
  ]
  \addplot+[thick] table[x expr=\thisrowno{1}/\epochs,x index={1},y index={3},col sep=comma, each nth point=\nthpoint, filter discard warning=false]{./tikz/results_base/GNN_Plot_data-HAWK_2025_02_17-run03-relexi_run_014/data.csv};
  \addplot+[thick,dashed] table[x expr=\thisrowno{1}/\epochs,x index={1},y index={3},col sep=comma, each nth point=\nthpoint, filter discard warning=false]{./tikz/results_base/GNN_Plot_data-HAWK_2025_02_17-run03-relexi_run_014/data_shuffled.csv};


  \nextgroupplot[
    ymin = -0.08, ymax = 0.08,
    xlabel={$t^*$},
    ylabel={$\hat{Q} / Q_{\text{ref}}$},
    scaled y ticks=false, 
    yticklabel style={/pgf/number format/fixed, /pgf/number format/precision=2}, 
  ]
  \addplot+[thick] table[x expr=\thisrowno{1}/\epochs,x index={1}, y expr=\thisrowno{4}, y index={4},col sep=comma, each nth point=\nthpoint, filter discard warning=false]{./tikz/results_base/MLP_Plot_data-flash_2025_02_27-run00-relexi_run_004/data.csv};
  \addplot+[thick,dashed] table[x expr=\thisrowno{1}/\epochs,x index={1}, y expr=\thisrowno{4}, y index={4},col sep=comma, each nth point=\nthpoint, filter discard warning=false]{./tikz/results_base/MLP_Plot_data-flash_2025_02_27-run00-relexi_run_004/data_shuffled.csv};

  \nextgroupplot[
    ymin = -0.08, ymax = 0.08,
    xlabel={$t^*$},
    yticklabels={,,},
    scaled y ticks=false, 
    yticklabel style={/pgf/number format/fixed, /pgf/number format/precision=2}, 
    legend pos=south east,
    legend cell align={left},
    legend style={font=\scriptsize},
  ]
  \addplot+[thick] table[x expr=\thisrowno{1}/\epochs,x index={1}, y expr=\thisrowno{4}, y index={4},col sep=comma, each nth point=\nthpoint, filter discard warning=false]{./tikz/results_base/GNN_Plot_data-_flash_2025_02_04-run02_relexi_run_010/data.csv};
  \addplot+[thick,dashed] table[x expr=\thisrowno{1}/\epochs,x index={1}, y expr=\thisrowno{4}, y index={4},col sep=comma, each nth point=\nthpoint, filter discard warning=false]{./tikz/results_base/GNN_Plot_data-_flash_2025_02_04-run02_relexi_run_010/data_shuffled.csv};

  \end{groupplot}
\end{tikzpicture}
    \caption{Time evolution of the lift ($C_L$) and drag ($C_D$) coefficients, and the normalized control outputs ($\hat{Q} / Q_{\text{ref}}$) for the MLP (\textit{left}) and GNN (\textit{right}) policy with either shuffled inputs or the original sorting of the probes as used during training.}%
    \label{fig:res:shuffled_input}
\end{figure}
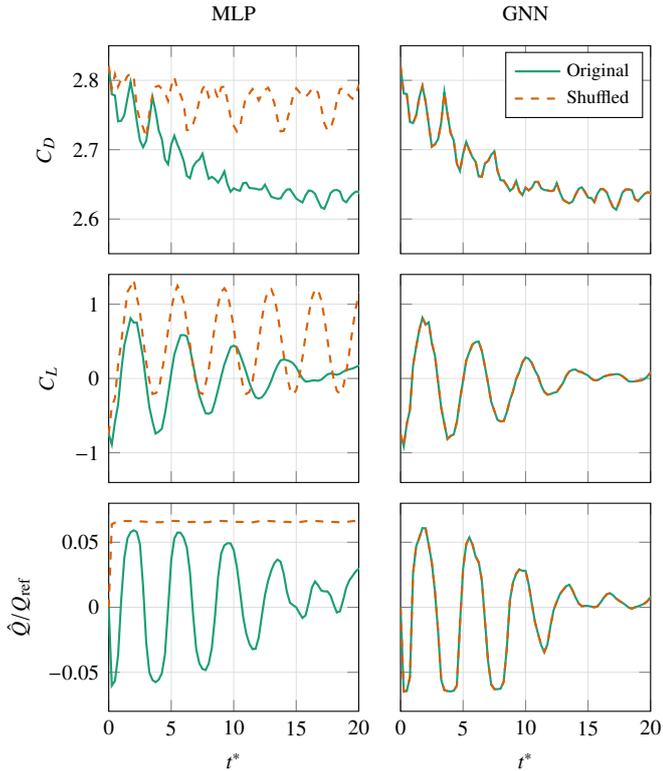

To verify the permutation invariance of the GNN policy and investigate the impact of the probe ordering on the MLP policy, we evaluate both policies for the same simulation but with shuffled inputs.
This means that the order in which the states of each probe are concatenated is changed in comparison to the original training data.
The resulting evolution of the lift and drag coefficients as well as the control output of the MLP and GNN policies are shown in \cref{fig:res:shuffled_input}.
As expected, the GNN policy is invariant to the ordering of the probes and yields the exact same results for the shuffled input as for the original input ordering.
This demonstrates the permutation invariance of the GNN policy.
In contrast, the MLP policy fails to yield any meaningful control output for the shuffled input and just produces the maximum control value during the simulation.
This again underlines the potential of GNNs for active flow control tasks, as they do not rely on an explicit ordering or flattening of the input data, but rather operate on the unstructured spatial structure of the data directly.

A key advantage of permutation invariance is that it exhibits a direct connection to more fundamental physical symmetries, such as rotational invariance.
This holds because, for scalar inputs, a rotation or reflection of the domain---or of the physical phenomena within it---is inherently captured through reshuffling, which encompasses all these operations as a superset.

While a GNN thus could be applied directly to a rotated or reflected domain without any need for retraining, an MLP would require a reordering of the input data to match the original training data.
The GNN policy thus promises to yield more robust and generalizable policies for more complex flow control tasks and to embed symmetries of the underlying physical phenomena directly into the policy.

\subsection{Analysis of Controlled Flow and Strategy}
\label{sec:results:strategy}
Both policy networks end up learning nuanced control strategies, like those reported by \citet{rabault2019artificial, jiang2023reinforcement} and \citet{jia2024deep}.
And much like these studies, the long-term controlled flow approximates the hypothetical no-shedding baseline flow in terms of the forces on the cylinder.
The superlative performance achieved from $t^*\approx 10$ to $t^* = 20$ (the simulation end time in the training phase) proves to be transient, and quickly asymptotes to a quasi-steady state that recovers the overall force $C_F$ of the symmetric case.
This could be due to the limited time duration of the simulations used during training and the resulting lack of future information in the latter stages of the training simulations.
For the first time steps of the simulations, the \textit{generalized advantage estimate} in the state-action-value tuple contains detailed information of the future and how actions taken at this moment affect the flow.
Based on this, the DRL method learns to take optimal steps keeping the effect it has on the future in mind.
Towards the end of the episode ($t^* \in [10,20]$) the agent has limited information of this context, since the simulation is halted at $t_{end}=20$ in every episode.
The agents thus only learn to take actions that have a positive effect on their immediate future, with no regard to what happens after the simulation ends.
This performance degradation of finite-horizon controllers in evaluations on longer time horizons is a well recognized issue in the field of optimal-control, and not just a RL issue \cite{sutton2020reinforcement, kaelbling1996reinforcement}.


\begin{figure}
  \centering
  \tikzsetnextfilename{fig_results_field_solution}
  \input{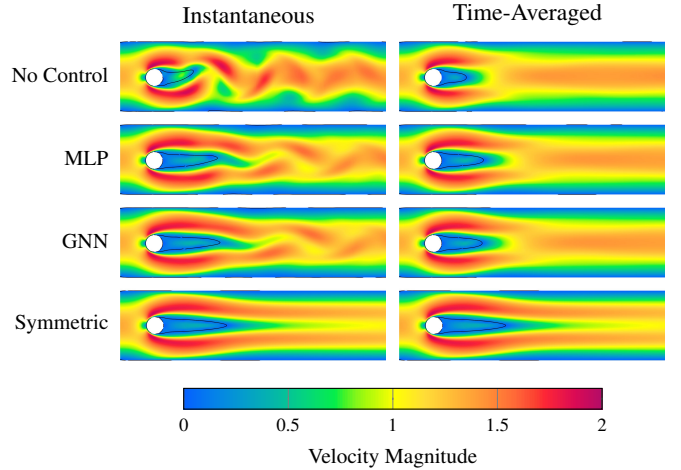}
  \caption{Instantaneous flow field at $t^*=20$ (\textit{left}) and time-averaged flow field in $t^*\in[50,100]$ (\textit{right}) for the no-control case, the GNN policy, the MLP policy, and the symmetric case colored by velocity magnitude. The black lines show the isocontour of vanishing velocity and thus the outline of the recirculation zone.}%
  \label{fig:res:field_solution}
\end{figure}

The instantaneous and time-averaged flow field solutions for the no-control case, the GNN and MLP policies, and the symmetric case are shown in \cref{fig:res:field_solution}.
Both the GNN and MLP policies are able to suppress the vortex shedding resulting in a noticeably longer recirculation bubble than in the uncontrolled case.
Interestingly, this recirculation bubble is larger for the instantaneous flow field at $t^*=20$ than for the time-averaged flow field, which directly reflects the lower drag observed during training (i.e. $t^*\leq 20$) than in the quasi-steady state as discussed prior.
Also, both the GNN and MLP policies recover similar drag reduction as the symmetric case while creating only a smaller recirculation zone.
This highlights again that the policies are able to find a more intricate control strategy than just suppressing the vortex shedding.

\section{Conclusions}
\label{sec:conclusion}
This study explored the feasibility of using GNNs for RL-based active flow control with a distinct focus on overcoming limitations associated with standard architectures such as MLPs and CNNs.
To the best of our knowledge, this is the first work to employ GNNs for active flow control.
The primary objectives were to develop a control strategy that efficiently processes unstructured flow data, while respecting the inherent invariance properties of the underlying physics to improve its robustness and generalization abilities.
By revisiting the well-established two-dimensional cylinder benchmark problem, the effectiveness of this approach was assessed and compared against existing methods.


The training setup was validated by training an MLP architecture similar to the original study by~\citet{rabault2019artificial}, which recovered similar performance as that reported in the reference.
The GNN policy network was found to successfully match the performance of this MLP architecture, confirming that GNNs can achieve similar levels of accuracy while offering additional generalization capabilities.
Both networks are able to reduce the drag below that observed in the symmetric case, where the vortex shedding is suppressed by imposing a symmetry condition along the centerline of the domain.
This indicates that the performance of both policies stems from more than just a reduction of the vortex shedding, and points to the effectiveness of the implemented PPO approach at finding intricate, well-performing strategies.
A key advantage of using GNNs over MLPs is the incorporation of permutation invariance into the control policy, possibly improving the model’s ability to generalize across different conditions.
This property was demonstrated successfully for the GNN, which yields identical results independent of how the input probes are ordered.
In contrast, the MLP was found to outright fail if the input ordering is changed.
This permutation invariance can be seen as a prerequisite for more general rotational and reflectional invariance, and is thus a distinguishing attribute in favor of the GNN architectures over the MLP.
To investigate the performance of the trained policies on longer time-scales, they were evaluated on simulations lasting $5$ times as long as those used for training.
The learned policies were found to converge to a quasi-steady behavior that matches the reduction in overall force observed in the symmetric case.
This work has demonstrated that Relexi is a suitable tool for flow control tasks and the combination with FLEXI as a flow solver positions it to be well-suited for large-scale flow-control applications in fluid dynamics.

\section*{Acknowledgements}
This work was partially funded by the European Union.
It has received funding from the European High Performance Computing Joint Undertaking (JU) and Sweden, Germany, Spain, Greece, and Denmark under grant agreement No 101093393.
Moreover, this research presented was funded by Deutsche For\-schungs\-ge\-mein\-schaft (DFG, German Research Foundation) under Germany's Excellence Strategy EXC 2075 -- 390740016, by the DFG Rebound -- 420603919, and in the framework of the research unit FOR 2895.
We acknowledge the support by the Stuttgart Center for Simulation Science (SimTech).
The authors also thank the Friedrich und Elisabeth Boysen-Stiftung for supporting the work under grant BOY187.
Furthermore, the authors gratefully acknowledge the support and the computing time on ”HAWK” provided by the HLRS through the project ”hpcdg”.
This work was carried out during the tenure of an ERCIM `Alain Bensoussan' Fellowship Programme by M.K.\ .

\section*{Data Availability Statement}
The Relexi and FLEXI codes used within this work are available under the \href{https://www.gnu.org/licenses/gpl-3.0.html}{GPLv3} license at:
\begin{itemize}
	\item \url{https://github.com/flexi-framework/relexi}
	\item \url{https://github.com/flexi-framework/flexi}
	\item \url{https://github.com/flexi-framework/flexi-extensions/tree/smartsim}
\end{itemize}
The implementation of the GNN in TensorFlow is available under the \href{https://mit-license.org/}{MIT} license at:
\begin{itemize}
  \item \url{https://github.com/m-kurz/gcnn}
\end{itemize}

The data generated in the context of this work and instructions to reproduce them with these codes are made available under the \href{https://creativecommons.org/licenses/by/4.0/}{CC-BY~4.0} license at:
\begin{itemize}
	\item \href{https://doi.org/10.18419/darus-4820}{{\tt 10.18419/darus-4820}}
\end{itemize}

\appendix

\section{Validation of the Simulation Setup}
\label{sec:validation}

This section provides a validation of the baseline simulation setup used in this study.
For this, the results obtained with FLEXI for the test case of a two-dimensional cylinder immersed in a box with viscous (no-slip) walls are compared against the reference~\cite{rabault2019artificial}.
The reader is referred to \citet{krais2021flexi} for details on the implementation of the discontinuous Galerkin approach and to \citet{hindenlang2014improving} for the validation of the convergence properties.
\begin{figure}[h]
  \tikzsetnextfilename{fig_validation_mach}
  \begin{tikzpicture}[font=\footnotesize, remember picture]

\def\XMIN{ 2.1}
\def\XMAX{10.2}
\def\YMIN{2.50}
\def\YMAX{3.7}
\def\DATAFILEONE{tikz/results_validation/Re100_Ma001_BodyForces_BC_wall_cylinder.csv}
\def\DATAFILETWO{tikz/results_validation/Re100_Ma010_BodyForces_BC_wall_cylinder.csv}
\def\DATAFILETHR{tikz/results_validation/Re100_Ma020_BodyForces_BC_wall_cylinder.csv}
\def\AXISSHIFTX{1cm}
\def\AXISSHIFTY{1cm}

\def\figwidth{1.00\linewidth}
\def\figheight{0.65\linewidth}

\def\thresholdX{3} 

\pgfplotsset{table/col sep=comma, table/header=true}

\begin{axis}[
    colormap/Reds,
    colormap/Blues,
    colormap/Greens,
    colormap/Oranges,
    width=\figwidth,
    height=\figheight,
    xlabel={Order $N$ of the polynomial ansatz},
    ylabel={$\langle C_D \rangle$},
    xmin=\XMIN, xmax=\XMAX,
    ymin=\YMIN, ymax=\YMAX,
    xticklabel style={/pgf/number format/fixed},
    yticklabel style={/pgf/number format/fixed},
    grid=both,
    grid style={line width=.1pt, draw=gray!5},
    major grid style={line width=.2pt,draw=gray!25},
    legend pos=north east,
    legend cell align={left},
    legend columns=2, transpose legend
]

\addplot[dashed, draw=black, domain={0:11}] {3.205};
\addlegendentry{Reference}

\def\hLevel{1}
\addplot[smooth,                 dashed, draw=mapped color, index of colormap=6 of Greens, forget plot] table [x expr = \thisrow{pLevel}, y expr = 2*\thisrow{xForce}, restrict expr to domain={\thisrow{hLevel}}{\hLevel:\hLevel}, restrict expr to domain={\thisrow{pLevel}}{ 0:\thresholdX}]{\DATAFILEONE};
\addplot[smooth, mark=triangle*, thick , draw=mapped color, index of colormap=6 of Greens             ] table [x expr = \thisrow{pLevel}, y expr = 2*\thisrow{xForce}, restrict expr to domain={\thisrow{hLevel}}{\hLevel:\hLevel}, restrict expr to domain={\thisrow{pLevel}}{\thresholdX:10}]{\DATAFILEONE};
  \addlegendentry{$\mathrm{Ma}=0.01$}

\def\hLevel{1}
\addplot[smooth,                 dashed, draw=mapped color, index of colormap=6 of Blues, forget plot] table [x expr = \thisrow{pLevel}, y expr = 2*\thisrow{xForce}, restrict expr to domain={\thisrow{hLevel}}{\hLevel:\hLevel}, restrict expr to domain={\thisrow{pLevel}}{ 0:\thresholdX}]{\DATAFILETWO};
\addplot[smooth, mark=square*,   thick , draw=mapped color, index of colormap=6 of Blues             ] table [x expr = \thisrow{pLevel}, y expr = 2*\thisrow{xForce}, restrict expr to domain={\thisrow{hLevel}}{\hLevel:\hLevel}, restrict expr to domain={\thisrow{pLevel}}{\thresholdX:10}]{\DATAFILETWO};
\addlegendentry{$\mathrm{Ma}=0.10$}

\def\hLevel{1}
\addplot[smooth,                  dashed, draw=mapped color, index of colormap=6 of Oranges, forget plot] table [x expr = \thisrow{pLevel}, y expr = 2*\thisrow{xForce}, restrict expr to domain={\thisrow{hLevel}}{\hLevel:\hLevel}, restrict expr to domain={\thisrow{pLevel}}{ 0:\thresholdX}]{\DATAFILETHR};
\addplot[smooth, mark=diamond*,   thick , draw=mapped color, index of colormap=6 of Oranges             ] table [x expr = \thisrow{pLevel}, y expr = 2*\thisrow{xForce}, restrict expr to domain={\thisrow{hLevel}}{\hLevel:\hLevel}, restrict expr to domain={\thisrow{pLevel}}{\thresholdX:10}]{\DATAFILETHR};
\addplot[        mark=otimes   , thick , draw=black, mark options={scale=1.5}           , forget plot] table [x expr = \thisrow{pLevel}, y expr = 2*\thisrow{xForce}, restrict expr to domain={\thisrow{hLevel}}{\hLevel:\hLevel}, restrict expr to domain={\thisrow{pLevel}}{       3.5:4.5}]{\DATAFILETHR};
\addlegendentry{$\mathrm{Ma}=0.20$}

\draw[-stealth, thick] (9.200, 2.75) -- (9.200, 3.10) node[midway, right] {$\mathrm{Ma}\!\downarrow$};

\node[inner sep=2pt] (annotation) at (axis cs:3.1, 2.57) {\scriptsize This Work};
\draw[-latex, thick] ($(annotation.north)!0.5!(annotation.north east)$) -- (axis cs:3.9, 2.72);

\end{axis}
\end{tikzpicture}
  \caption{Convergence of time-averaged drag coefficient $\langle C_D\rangle$ against order of the polynomial ansatz for different Mach numbers. The results by \citet{rabault2019artificial} serve as reference. The selected setup for this work is highlighted in black.}%
  \label{fig:validation:mach}
\end{figure}
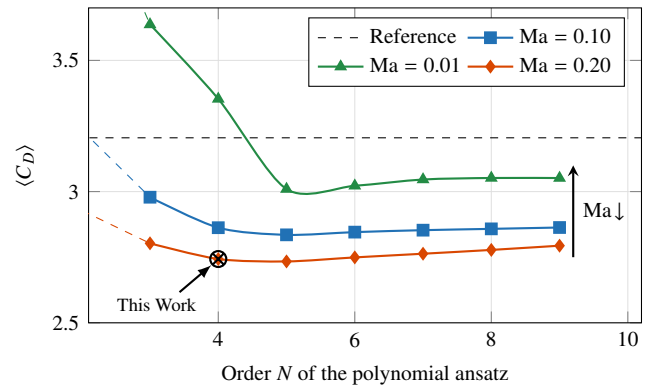%
The resulting average drag coefficient for increasing orders $N$ of the polynomial ansatz function are show in \cref{fig:validation:mach} for three different Mach numbers $\mathrm{Ma} = \{0.01, 0.1, 0.2\}$.
All considered Mach numbers exhibit a marked increase in drag for very low polynomial orders, followed by a local minimum of $\langle C_D\rangle$ at approximately $N \approx 4$ and subsequent convergence. Furthermore, the setup is shown to react sensitively to the free-stream Mach number even as the results by FLEXI converge towards the incompressible reference for decreasing Mach numbers. A plausible explanation of the reduced drag is the interaction of the limited domain size with the weakly enforced Dirichlet boundary conditions. This interplay results in a mass-flow deficit directly related to the $\langle C_D\rangle$ difference, see \cref{tab:validation:mass}. From these results, the $N=4$, $\mathrm{Ma}=0.2$ case (highlighted in \cref{fig:validation:mach}) was chosen as a good compromise between computational effort and accuracy.

\begin{table}[h]
  \normalsize
  \begin{tabular*}{\linewidth}{@{\extracolsep{\fill}} lcc}
    \toprule%
    \textbf{Mach number $\mathrm{Ma}$} & $\Delta \dot{m}$ & $\Delta \langle C_D\rangle$\\
    \midrule%
    $0.2$                     & $0.0943$         & $0.1282$\\
    $0.1$                     & $0.0722$         & $0.1066$\\
    $0.01$                    & $0.0366$         & $0.0478$\\
    \bottomrule%
  \end{tabular*}
  \caption{Mass-flow $\dot{m}$ and drag coefficient $\langle C_D\rangle$ differences from the target values for the $N=9$ cases.}%
  \label{tab:validation:mass}
\end{table}

\section{Symmetric Simulation Setup}%
\label{sec:symmetric}

\begin{figure}[h]
	\centering
	\tikzsetnextfilename{fig_baseline_appendix}
	\input{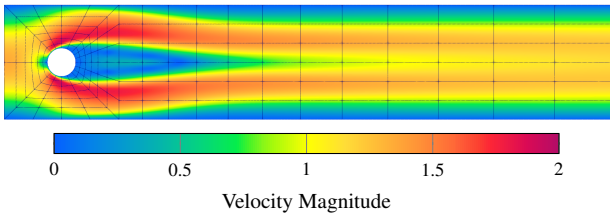}
	\caption{Mesh and converged flow field of the symmetric setup. Note the lack of vortex shedding in the wake of the cylinder.}%
	\label{fig:sym:mesh_flowfield}
\end{figure}

We also simulate a hypothetical no-vortex-shedding case that serves as an ideal flow case with minimum drag that the AFC should be able to achieve.
This allows us to compare the achieved drag reduction with the reference studies in light of the compressibility effects highlighted in \cref{sec:validation}.
The setup follows \cite{rabault2019artificial}, where the cylinder is placed in the center of the channel and a symmetry condition is imposed at the centerline that forces the normal gradient and the normal component of the velocity vectors to zero.
This simulation setup is advanced up to $t^*=500$ to reach a convergened state.
The results reported in \cref{sec:results} of the main text are obtained by restarting the simulation from this converged state.
The mesh and flow field of the converged simulation are shown in \cref{fig:sym:mesh_flowfield}.

\bibliographystyle{elsarticle-harv}
\bibliography{references.bib}



\end{document}